%% file: acl.tex
\pdfoutput=1

\documentclass[11pt]{article}

\usepackage[]{acl}

\usepackage{times}
\usepackage{latexsym}
\usepackage{booktabs}
\usepackage{algorithm}
\usepackage{algorithmic}

\usepackage{amsmath}
\usepackage{multirow}
\usepackage{adjustbox}
\usepackage{lipsum}
\usepackage{xspace}
\usepackage{makecell}
\usepackage{color}
\usepackage{arydshln}
\usepackage{xspace}

\usepackage{pifont}
\usepackage{url}
\newcommand{\cmark}{\ding{51}}
\newcommand{\xmark}{\ding{55}}
\newcommand{\ours}{\textsc{Summ$^N$}\xspace}

\newcommand{\green}[1]{\textcolor{brown}{#1}}
\newcommand{\red}[1]{\textcolor{red}{#1}}
\newcommand{\gray}[1]{\textcolor{gray}{#1}}
\usepackage[T1]{fontenc}
\usepackage{graphicx}
\usepackage[utf8]{inputenc}

\usepackage{microtype}

\title{\ours: A Multi-Stage Summarization Framework for Long Input Dialogues and Documents}

\author{
Yusen Zhang$^\clubsuit$ 
\quad Ansong Ni$^\dagger$
\quad Ziming Mao$^\dagger$
\quad Chen Henry Wu  $^\ddagger$

\\{\bf 
\quad Chenguang Zhu$^\diamondsuit$
\quad Budhaditya Deb$^\diamondsuit$
\quad Ahmed H. Awadallah$^\diamondsuit$}

\\{\bf 
\quad Dragomir Radev$^\dagger$
\quad Rui Zhang$^\clubsuit$ 
} 

\\
$^\clubsuit$ Penn State University
\quad $^\dagger$ Yale University 
\\
\quad $^\ddagger$ Carnegie Mellon University
\quad $^\diamondsuit$ Microsoft Research
\\
\tt{\{yfz5488,rmz5227\}@psu.edu},
\tt{\{ansong.ni,dragomir.radev\}@yale.edu}
}

\begin{document}
\maketitle
\begin{abstract}
Text summarization helps readers capture salient information from documents, news, interviews, and meetings. However, most state-of-the-art pretrained language models (LM) are unable to efficiently process long text for many summarization tasks. In this paper, we propose \ours, a simple, flexible, and effective multi-stage framework for input texts that are longer than the maximum context length of typical pretrained LMs. \ours first splits the data samples and generates a coarse summary in multiple stages and then produces the final fine-grained summary based on it. Our framework can process input text of arbitrary length by adjusting the number of stages, while keeping the LM input size fixed. Moreover, it can deal with both single-source documents and dialogues, and it can be used on top of different backbone abstractive summarization models. To the best of our knowledge, \ours is the first multi-stage split-then-summarize framework for long input summarization. Our experiments demonstrate that \ours outperforms previous state-of-the-art methods by improving ROUGE scores on three long meeting summarization datasets AMI, ICSI, and QMSum, two long TV series datasets from SummScreen, and a long document summarization dataset GovReport. Our data and code are available at \url{ https://github.com/psunlpgroup/Summ-N}. 
\end{abstract}

\begin{figure*}[ht!]
    \centering
    \includegraphics[width=\textwidth]{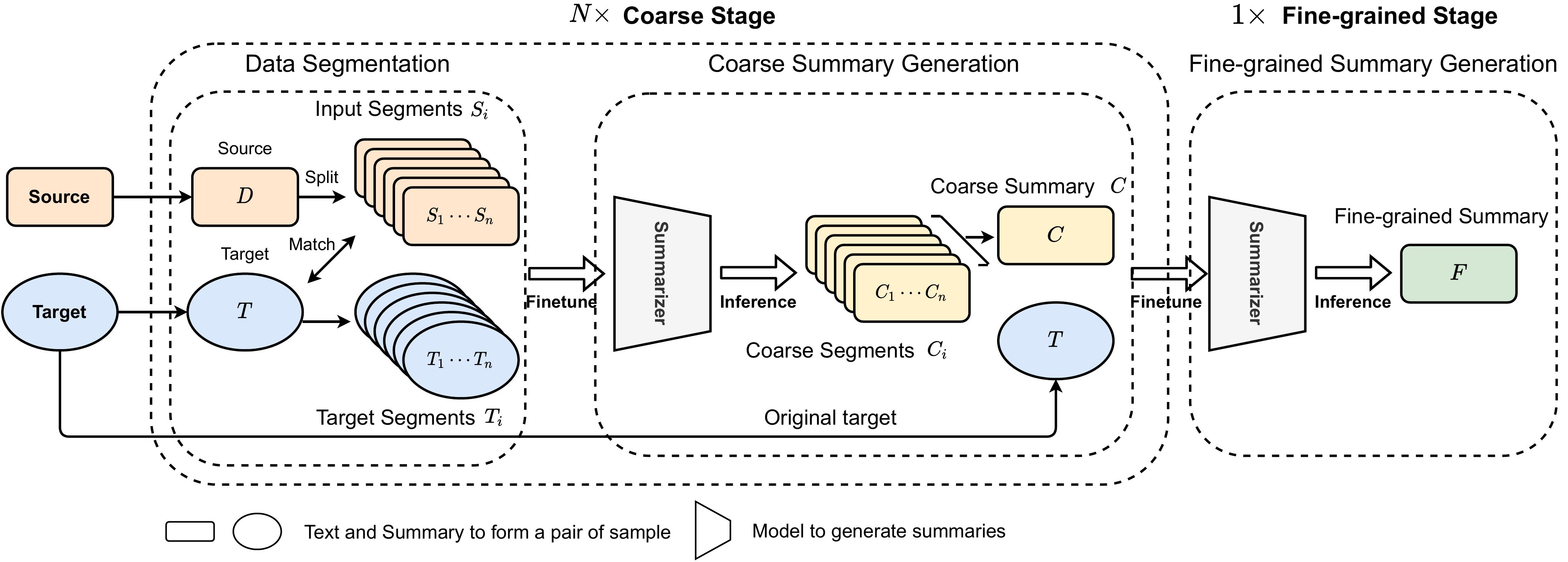}
    \caption{Workflow of the proposed \ours framework. It contains $N$ coarse stages and $1$ fine-grained stage. At each coarse stage, source and target text is segmented and paired using a ROUGE-based greedy algorithm, and then a backbone summarization model is used to generate the summary for each segment. After multiple coarse stages, the last fine-graded stage produces the final summary output. 
    }
    \label{fig:input}
\end{figure*}

\section{Introduction}
Abstractive summarization helps readers capture salient information from various sources such as documents, news, interviews, and meetings. Previous work has primarily focused on short texts of news~\citep{gehrmann-etal-2018-bottom, zhang2019pegasus} and short conversations~\citep{gliwa-etal-2019-samsum, chen-yang-2021-structure}.
Recently proposed longer dialogue and document summarization tasks~\citep{zhong2021qmsum, HuangCPJW21, chen2021summscreen, zhu-etal-2021-mediasum} pose challenges for current large pretrained language models due to the time and memory complexity of training, as well as limited input lengths these models can consume.

A common method to handle long text reduces the input to a shorter one. This can be accomplished by truncating inputs~\citep{lewis-etal-2020-bart} or employing retrieve-then-summarize pipelines~\citep{zhong2021qmsum}.
However, these methods break the dependency of the context and decrease the number of tokens that the model can read, i.e., the receptive field of the model. The cutting-off model depends on the lead bias of the source text, while the retrieve-then-summarize models heavily rely on the independence of retrieved units (turns or sentences) which are usually scattered throughout the source text.

Another approach optimizes the attention mechanism in Transformers to accommodate longer inputs by reducing the impact of quadratic complexity of the attention process using Locality-sensitive hashing (LSH) attention~\citep{kitaev2020reformer} and Sinkhorn attention~\citep{tay2020sparse}. Additionally, HMNet~\citep{zhu-etal-2020-hierarchical} and HAT-BART~\citep{rohde2021hierarchical} use hierarchical self-attention to extend the input limitation of typical self-attention models. However, the simplified attention mechanism weakens the power of pretrained Transformer models, e.g., HMNet is not pretrained on external large-scaled unsupervised datasets as BART did.

In this paper, we propose \ours, a multi-stage framework for long dialogue and document summarization. Figure~\ref{fig:input} shows the structure of \ours. First, it divides each source text into segments so that each can be completely fed into the backbone abstractive summarization model. Then, it matches each of them with the subset of target text using a ROUGE-based greedy algorithm. 
Next, each stage generates a coarse summary for each segment and concatenates them together as the input to the next stage. After multiple stages of compression and summarization, the final stage produces a fine-grained summary. The process expands the model context to the full reception field, meaning that the proposed model can read the full input no matter how long the input is. Additionally, retrieve-then-summarize pipelines~\citep{zhang2019pegasus} extract sentences individually, leading to the loss of the context information for understanding utterances. By contrast, \ours only cuts the source text at the end of each segment, so that the context of most sentences are retained.

It does not assume lead bias because each part of the source is fully used. In addition, in each stage, it leverages a backbone abstractive summarization model to recursively generate the summaries. Therefore, it enjoys the full power of the pretrained language models because the framework preserves the intact structure of Transformers.

\ours is flexible to inputs with different lengths by adjusting the number of stages.
\ours can change the number of coarse stages according to the compression ratio between source and target, the input limit of the backbone model, and the input source length. We give the empirical formula to decide the number of needed stages for every tested dataset. Our experiments show that ROUGE increases on all datasets when increasing the number of stages from one to the appropriate number.
Additionally, \ours is flexible because it can be applied to different backbone summarization models.
For example, we found that the ROUGE scores increase sharply on the AMI dataset when replacing the backbone BART model with T5~\citep{2020t5} and PEGASUS~\citep{zhang2019pegasus}.

We conduct extensive experiments on long-input summarization datasets in multiple domains. The results demonstrate that the proposed model significantly outperforms previous state-of-the-art methods according to automatic and human evaluations on three long meeting summarization datasets (AMI, ICSI, QMSum) and one long TV series summarization dataset (SummScreen). It also achieves state-of-the-art performance on a long document summarization dataset (GovReport). These datasets include document summarization as well as both query-based and query-independent long dialogue summarization tasks.

Our contributions are: (1) We propose \ours, a simple, flexible, and effective framework for long dialogue and document summarization. To the best of our knowledge, \ours is the first multi-stage split-then-summarize framework to solve long text summarization tasks. (2) We evaluate \ours on both dialogue and document domains and improve the baseline model by a large margin. (3) We analyze and compare the proposed framework with baselines and discuss its merits in detail.

\section{Related Work}
\paragraph{Long Document Summarization}
Long document summarization has been studied in multiple domains, such as news~\citep{liu2021end,zhu2021leveraging}, patterns~\citep{trappey2009automatic}, books~\citep{kryscinski2021booksum, wu2021recursively}, scientific publications~\citep{qazvinian-radev-2008-scientific,mao2021dyle}, and medical records~\citep{cohan-etal-2018-discourse}. 
\citet{9257174} proposed a divide-and-conquer method by splitting the input into multiple segments, summarizing them separately, and combining the summary pieces. \citet{grail-etal-2021-globalizing} proposed a hierarchical neural model to process segmented input blocks. Compared with \ours, these models only split the input once, implying the lack of flexibility when handling longer input.

The GovReport dataset was recently introduced containing documents with more than 9000 words, thus greatly challenging the capabilities of current models such as PEGASUS~\citep{zhang2019pegasus}, TLM~\citep{subramanian2019extractive}, and BIGBIRD~\citep{zaheer2020big}. To handle this dataset,
\citet{HuangCPJW21} proposed head-wise positional strides to reduce the cost of the encoder-decoder attention. Similarly, models such as Longformer~\citep{Beltagy2020Longformer} and Reformer~\citep{kitaev2020reformer} adjust attention mechanisms in Transformers to consume longer inputs.
However, these models sparsify the attention structure of the pretrained model to fit the longer source text. By contrast, \ours is able to maintain the full structure of various pretrained models.

\paragraph{Long Dialogue Summarization}
Various models have also been proposed to handle long dialogue summarization.
HMNet~\citep{zhu-etal-2020-hierarchical} and HAT-BART~\citep{rohde2021hierarchical} leverage a two-level transformer-based model to obtain word level and sentence level representations. DialLM~\citep{zhong2021dialoglm}, Longformer-BART-arg~\citep{fabbri-etal-2021-convosumm} use finetuning or data augmentation to incorporate the external knowledge to maintain the accuracy of lengthy input. Different from these models, \ours is a framework without modifying the structure of the backbone attention model.

\paragraph{Multi-Stage Text Generation}
Multiple multi-stage coarse-to-fine frameworks have been studied in many other text generation tasks, such as dialogue state tracking~\citep{chen2020credit}, neural story generation~\citep{fan-etal-2018-hierarchical}, and extractive summarization~\citep{xu-lapata-2020-coarse}. In a summarization task, a two-stage extract-and-summarize pipeline is commonly used~\citep{zhang2019pegasus, subramanian2019extractive, zhao2020seal}. However, unlike that work, our framework aims at long input summarization with fully abstractive intermediate summaries, meaning that \ours can be viewed as a summarize-then-summarize pipeline.

\section{Method}
Figure \ref{fig:input} shows the workflow of \ours. The workflow includes two types of stages, $N$ coarse stages, and one fine-grained stage. Coarse stages include the data segmentation and coarse summary generation, while the fine-grained stage directly generates the summary as the final result. Besides, we have $N+1$ separate models for each stage and each was separately trained. Our experiments show that the performance drops if different stages share the parameters (Section \ref{sec:backbone}). \ours can adjust and compute the number of coarse stages $N$ according to the stats of dataset and model.

To formulate our task, we denote one sample of the source text as $D = \{D_1, D_2, \cdots, D_m\}$, where $D_i$ indicates one sentence in a document or one turn in a dialogue. For query-based summarization, there is also a query $Q$. The goal is to generate a summary $T$, given $D$ and the optional $Q$.

\subsection{Data Segmentation} 
In long text summarization, the number of tokens in the source data usually exceeds the limit of the backbone summarization models, thus reducing the quality of the summary. To make sure that the model can capture information about all source tokens, we apply a segmentation algorithm for long input summarization datasets. First, we segment the source text so that the data input to the backbone model does not exceed the length limit. Then, we apply a greedy algorithm to find the best target summary that matches the source segments.

\paragraph{Source Segmentation} Assume that the number of the maximum input tokens of the backbone model is $K$. To completely input the source information, we cut the input $D$ (between sentences) into multiple segments, each of them containing fewer than $K$ tokens. Given the input $D$, we will have $n$ segments $S = \{S_1, S_2, \cdots, S_n\}$ where $S_i \in D$ is a segment in $D$. For query-based summarization tasks, we simply concatenate the query to the beginning of the $S$, i.e. $ S_i \gets Q \bigoplus S_i$. In both cases, the number of tokens in each segment is less than the hyper-parameter $K$.

\begin{algorithm}[t!]
\small
	\caption{Greedy Target Matching} 
	\label{alg1} 
	\begin{algorithmic}
		\REQUIRE $S_i$, $T_s = \{T_{s_1}, T_{s_2}, \cdots, T_{s_k}\}$
		\ENSURE $(S_i, T_i)$ 
		\STATE $T_i \gets \Phi$

		\LOOP
		\STATE $T'_i \gets T_i$
		\FOR {$T'_s \in T_s - T_i$ }
		    \STATE $\tau' \gets \text{ROUGE}_1(S_i, T'_i)$
		    \STATE $\tau \gets \text{ROUGE}_1(S_i, T_i \bigoplus T'_s)$
		    \IF {$ \tau' < \tau $} 
                \STATE $T'_i \gets T_i \bigoplus T'_s$
            \ENDIF
		\ENDFOR
		\IF {$T'_i = T_i$}
		\STATE Break the loop.
		\ELSE
		\STATE $T_i \gets T'_i$
		\ENDIF
		\ENDLOOP
		\RETURN $(S_i, T_i)$
	\end{algorithmic}
\end{algorithm}

\paragraph{Target Matching} 
Segmenting the source text results in $n$ source pieces $S_i$. We match each $S_i$ with a target segment $T_i \in T$ to form the new pair $(S_i, T_i)$ for the next step. We use a greedy algorithm for target matching.
We first split $T$ into separate sentences $T_s = \{T_{s_1}, T_{s_2}, \cdots, T_{s_k}\}$. Then, each segment $S_i$ is matched with a subset of $T_s$ such that the ROUGE-1 score between the subset and $S_i$ is maximized. However, it is not feasible to find the optimal set due to the considerable running time. We apply a simple greedy approximation to find such a subset. From a null set $T_i$, we iteratively add to the subset the sentence with the highest ROUGE-1 gain between $T_i$ and $S_i$. Algorithm \ref{alg1} shows how we obtain the new training pair $(S_i, T_i)$. $\bigoplus$ indicates the concatenation of sentences while keeping them in the same order as in the original text. We use ROUGE-1 as the matching criterion because the higher ROUGE-1 score usually implies higher scores on the other metrics such as ROUGE-2 or ROUGE-L, while ROUGE-1 enjoys lower time complexity compared with other ROUGE metrics.

This matching algorithm also ensures $T_i \neq \emptyset$ so that each $S_i$ can be matched to at least one target sentence. A sentence $t \in T_s$ can be added to multiple subsets $T_i$ because one sentence of summary may need the information from multiple segments.

\subsection{Coarse Summary Generation} 
In coarse summary generation, we train a summarization model, that takes the segmented data as input. We first collect the training samples $(S_i, T_i)$ generated by data segmentation to form a new dataset. This augments the source data to $d_1/K$ times compared with the cut-off methods, where $d_1 = |D^1|$ indicates the averaged number of tokens of original source text. Thus, data segmentation helps the summarizer to better learn the task of the current stage. Additionally, because we incorporate the full input using segmentation, it does not rely on the leading bias in the cut-off method that only considers the first segment $S_1$.
Afterward, we use these data to train a neural summarizer.
This way, our model treats each part of the source text as equally important.

Given a source segment $S_i$ and an optional query $Q$, we obtain the coarse summary segments using a backbone summarization model:
$$C_{i}^l = \text{SUMM}_l(Q,S_i)$$
Where $l \in [1,N]$ is the index of the current stage. Then, the $n$ coarse summaries corresponding to the original source $S=\{S_1, S_2, \cdots, S_n\}$ are concatenated: $C^l = C^l_1\bigoplus C^l_2\bigoplus \cdots\bigoplus C^l_n$. We use $C^l$ as the new source text of next stage, which compresses the input source data $D^{l}$. i.e. $D^{l+1} = C^{l}$. To pair with the $D^{l+1}$, the target to the next stage is copied from the original dataset, i.e. $T^{l+1} = T$. 

The proposed framework is applicable to different backbone models $\text{SUMM}_l(*)$, such as BART~\citep{lewis-etal-2020-bart} and T5~\citep{2020t5}. We pick BART as the backbone model because it can best illustrate the benefits of our framework (Section \ref{sec:backbone}).

\begin{table*}[ht!]
    \centering
    \resizebox{.85\textwidth}{!}{
    \begin{tabular}{lllrrrcc}
    \toprule
        Dataset & Type & Domain & Size & Source length & Target length & Query & $N + 1$ \\\midrule
        AMI & Dialogue & Meetings & 137 & 6007.7 &  296.6 & \xmark & 2 \\
        ICSI & Dialogue & Meetings & 59 & 13317.3 & 488.5 & \xmark & 3 \\
        QMSum & Dialogue & Meetings & 1808 & 9069.8 & 69.6 & \cmark & 2 \\
        SummScreen & Dialogue & TV shows & 26851 & 6612.5 & 337.4 & \xmark & 2 \\
        GovReport & Document & Reports & 19466 & 9409.4 & 553.4 & \xmark & 3\\
    \bottomrule
    \end{tabular}
    }
    \caption{The summarization datasets for evaluation. The source length and target length is the averaged number across the dataset. $N$ indicates the number of coarse stages we use. 
    } 
    \label{tab:datasets}
\end{table*}

\subsection{Estimation of the Number of Coarse Stages $N$}
\label{sec:corstage}
The number of stages can be estimated by data stats and model characteristics.
In \ours, each coarse stage compresses the input to a shorter length.
After $N$ turns of coarse stages, the averaged length of source text is below $K$, the dataset is then fed into the fine-grained stage. Hence, the number of coarse stages can be computed by the following equation (details can be found in Appendix A):
    $$ \hat{N} = \lceil \frac{\log K - \log d_1}{\log c_1 - \log K} \rceil$$
where $d_1$ and $c_1$ are the average length of source text and coarse segments in stage 1. In Section \ref{sec:intermedi} and Table \ref{tab:emp}, we demonstrate this estimation is close to the empirical number of coarse stages.

The greedy algorithm in \ours for target matching is critical to the performance. Consider a duplication algorithm where each segment $S_i$ is simply paired with the target $T$, i.e. $T_i = T$. Since the target text is longer than the text segmented by Algorithm \ref{alg1}, the generated summary of each coarse stage will be longer as well, leading to a lower compression speed and larger $N$. Besides, the duplication of the target will confuse the model, because some source segments will probably be paired with the same target, causing the model to generate duplicated content. Experiments (Table \ref{tab:ablation}, ``- stage 2'' versus ``- stage 2 \& tar. seg.'') show that ROUGE scores declines a lot when greedy target segment is replaced by the duplication algorithm .

\subsection{Fine-Grained Summary Generation}
When the input source of $D^l$ is shorter than $K$, we can proceed to the fine-grained stage. In this stage, $D^l$ is used to train a summarization model from scratch to obtain the final summary.
The fine-grained stage works the same way as the vanilla backbone model. In fact, \ours with $N = 0$ is the backbone summarizer.
In the fine-grained stage, the model is directly trained on dataset $(D^{N}, T)$ from the last coarse stage, and obtain the summary as the final output of \ours:
$$F = \text{SUMM}_{N+1}(Q,D^{N})$$ 

 It is worth noting that, although source text may be shorter than 2 segments, i.e. $d^i \leq K$, we still add them in all stages, so that each summarization model can be trained on the full dataset.
 
\section{Experiment Setup}
We first list the datasets and metrics to evaluate the model. Then, we introduce the backbone model and baselines for comparisons. Finally, we present some implementation details.

\subsection{Datasets and Metrics}
Table~\ref{tab:datasets} shows data statistics for the datasets\footnote{Both QMSum and SummScreen can be accessed through SummerTime \cite{ni-etal-2021-summertime}.}.

\paragraph{AMI \& ICSI}~\citep{mccowan2005ami,janin2003icsi} are meeting scripts generated by Automatic Speech Recognition (ASR) systems. AMI is collected from product design meetings in a company while ICSI is collected from academic group meetings. Because the transcript is produced by  ASR, there is a word error rate of 36\% for AMI and 37\% for ICSI.

\paragraph{QMSum}~\citep{zhong2021qmsum} is a query-based meeting summarization dataset. It consists of meetings from three domains, including AMI and ICSI, and the committee meetings of the Welsh Parliament and the Parliament of Canada. Each query and sample are written by experts. 

\paragraph{SummScreen}~\citep{chen2021summscreen} consists of community-contributed transcripts of television show episodes from The TVMegaSite, Inc. (TMS) and ForeverDream (FD). The summary of each transcript is the recap from TMS, or a recap of the FD shows from Wikipedia and TVMaze. 

\paragraph{GovReport}~\citep{HuangCPJW21} is a large-scale long document summarization dataset with 19,466 long reports published by the U.S. Government Accountability Office on national policy issues.

We use ROUGE~\citep{lin-2004-rouge} as the automatic evaluation metric.\footnote{We use \texttt{pyrouge},  a Python wrapper for the ROUGE: \\ https://github.com/bheinzerling/pyrouge}
We split summary outputs into sentences to calculate the ROUGE-L score.
If not specified, F1 scores are used in all results.

\begin{table*}[!ht]
\resizebox{\textwidth}{!}{
\begin{tabular}{@{}lrrrrrrrrrrrr@{}}
\toprule
            & \multicolumn{3}{c}{AMI} & \multicolumn{3}{c}{ICSI} & \multicolumn{3}{c}{QMSum-All} & \multicolumn{3}{c}{QMSum-Gold} \\ 
            & \multicolumn{1}{c}{R-1} & \multicolumn{1}{c}{R-2} & \multicolumn{1}{c}{R-L} & \multicolumn{1}{c}{R-1} & \multicolumn{1}{c}{R-2} & \multicolumn{1}{c}{R-L} & \multicolumn{1}{c}{R-1} & \multicolumn{1}{c}{R-2} & \multicolumn{1}{c}{R-L} & \multicolumn{1}{c}{R-1} & \multicolumn{1}{c}{R-2} & \multicolumn{1}{c}{R-L} \\ \midrule
PGNet       & 42.60  & 14.01  & 22.62*  & 35.89   & 6.92   & 15.67*  & 28.74     & 5.98    & 25.13    & 31.52    & 8.69     & 27.63    \\
TopicSeg    & 51.53  & 12.23  & 25.47*  & -       & -      & -      & -         & -       & -        & -        & -        & -        \\
HMNET       & 52.36  & 18.63  & 24.00*  & \textbf{45.97}   & 10.14  & 18.54*  & 32.29     & 8.67    & 28.17    & 36.06    & 11.36    & 31.27   \\
TextRank    & 35.19  & 6.13   & 16.70*   & 30.72   & 4.69   & 12.97*  & 16.27     & 2.69    & 15.41    & -        & -        & -        \\
HAT-BART    & 52.27  & 20.15  & 50.57  & 43.98   & 10.83  & 41.36  & -         & -       & -        & -        & -        & -        \\
DDAMS       & 53.15  & \textbf{22.32}  & 25.67*  & 40.41   & 11.02  & 19.18*  & -         & -       & -        & -        & -        & -        \\
\midrule

\ours & \textbf{53.44}  & 20.30   & \textbf{51.39} & 45.57 & \textbf{11.49} & \textbf{43.32} & \textbf{34.03}     & \textbf{9.28}    & \textbf{29.48}  & \textbf{40.20}     & \textbf{15.32}    & \textbf{35.62}    \\ 
\bottomrule
\end{tabular}
}
\caption{ROUGE scores on three meeting summarizing tasks, AMI, ICSI, and QMSum. QMSum-ALL uses inputs with all turns while MSum-Gold uses inputs with only the gold turns.
* denote the ROUGE-L scores without sentence split.}
\label{tab:meeting}
\end{table*}

\subsection{Backbone Model}
\label{sec:backbone}
We pick BART~\citep{lewis-etal-2020-bart} as our backbone summarization model because it performs well on short text summarization but not as good on longer texts, illustrating the benefits of our framework. Compared with other pretrained parameters, the BART-large model pretrained on the CNN/DM dataset yields the best performance~\citep{zhang2021exploratory}. So we use the BART-large-cnn parameter as a better starting point.

It is worth noting that we use separate backbone models for each stage and each was separately trained. We experimented with reusing the model parameters in multiple stages but obtained a lower score, e.g. the ROUGE-1 score of stage 2 on the QMSum dataset decreases around two points if we use the best parameters of stage 1 summarizer as the starting point of training stage 2 summarizer. This is because the tasks of the different stages differ significantly. For instance, the input to the first stage of dialogue summarization is the dialogue turn while the input to the latter stages is the document. 

\subsection{Baselines}
We compare the proposed framework with various baselines. \textbf{PGNet}~\citep{see-etal-2017-get} uses a pointer mechanism to copy the token from the training sample. \textbf{TopicSeg}~\citep{li-etal-2019-keep} is a multi-modal model jointly learning the segmentation and summarization. \textbf{HMNet}~\citep{zhu-etal-2020-hierarchical} uses a hierarchical attention structure and cross-domain pre-training for meeting summarization. \textbf{TextRank}~\citep{mihalcea-tarau-2004-textrank} is  a graph-based ranking model for text processing. \textbf{HAT-BART}~\citep{rohde2021hierarchical} is a new hierarchical attention transformer-based architecture that outperforms standard Transformers. \textbf{DDAMS}~\citep{DBLP:conf/ijcai/FengF0G21} uses a relational graph to model the interaction between utterances by modeling different discourse relations. 

For the SummScreen dataset, we use the neural and hybrid model scores reported by \citet{chen2021summscreen}. We rename these two baselines as \textbf{Longformer+ATT} and \textbf{NN+BM25+Neural} to clarify the difference between other baselines.

The baseline scores we report on GovReport are from the original paper~\citep{HuangCPJW21}. \textbf{BART Variant} indicates self-attention variants with full attention. \textbf{BART HEPOS} indicates encoder variants with head-wise positional strides (HEPOS) encoder-decoder attention.

\subsection{Implementation Details}

We fit all models into a single RTX A6000 GPU with a 48 GiB memory. We adopt the fairseq\footnote{https://github.com/pytorch/fairseq} implementation for BART. The learning rate is set to 2e-5 and the beam width is set to 2 for coarse stages and 10 for fine-grained stages. The maximum number of tokens in each batch is set to 2048. The maximum number of tokens in each source text is set to 1024 because we tried to extend the positional embeddings to 2048 or longer but obtained worse performance. We stop the coarse stage and start the fine-grained stage when the averaged source length is shorter than 2048 rather than 1024 to obtain a better performance (Section \ref{sec:intermedi}). For the output of each intermediate stage, we use <s> and </s> to separate each generated target segments $C^l_i$.

\section{Results and Analysis}
We discuss the evaluation results and effects of each component of \ours in this section.

\subsection{Overall Results}
\paragraph{Meeting Summarization} Table~\ref{tab:meeting} shows the ROUGE scores on AMI, ICSI, and QMSum. Compared with the baseline models, \ours achieves state-of-the-art results on almost all metrics. Specifically, \ours improves SOTA on ICSI by \textbf{0.83}, and \textbf{1.96} ROUGE-2/L scores, improves SOTA on QMSum-Gold by \textbf{4.14}, \textbf{3.96}, and \textbf{4.35} ROUGE-1/2/L scores. These results demonstrate the effectiveness of \ours on long dialogue summarization tasks.

\paragraph{TV Series Summarization} Table~\ref{tab:summscreen} shows ROUGE scores on SummScreen. \ours outperforms on almost all metrics on two SummScreen datasets. Specifically, we improve \textbf{6.58}, \textbf{1.65}, and \textbf{3.75} ROUGE-1/2/L scores on the SummScreen-FD dataset. This result demonstrates the generalizability of \ours over various domains including meetings and TV series.

\paragraph{Document Summarization} Table~\ref{tab:GovReport} shows ROUGE scores on GoveReport. \ours achieves state-of-the-art performance on ROUGE-2 and ROUGE-L, and compatible results on ROUGE-1. The results show that \ours is applicable to both long dialogue and document summarization tasks.

\begin{table}[!t]
\centering
\resizebox{\linewidth}{!}{
\begin{tabular}{@{}lrrrrrr@{}}

\toprule
            & \multicolumn{3}{c}{SummScreen-FD}              & \multicolumn{3}{c}{SummScreen-TMS} \\ 
            & R1             & R2            & R-L             & R1         & R2         & R-L         \\ \midrule
Longformer+ATT  & 25.90          & 4.20          & 23.80          & 42.90      & \textbf{11.90}      & 41.60      \\
NN+BM25+Neural      & 25.30          & 3.90          & 23.10          & 38.80      & 10.20      & 36.90      \\ \midrule
\ours & \textbf{32.48} & \textbf{5.85} & \textbf{27.55} & \textbf{44.64}  & 11.87  & \textbf{42.53} \\ \bottomrule
\end{tabular}
}
\caption{ROUGE scores on the SummScreen datasets including ForeverDreaming (SummScreen-FD) and TV MegaSite, Inc. (SummScreen-TMS).}
\label{tab:summscreen}
\end{table}

\begin{table}[t!]
\centering
\resizebox{.8\columnwidth}{!}{
\begin{tabular}{@{}lrrr@{}}
\toprule
~  			      & R-1 & R-2 & R-L \\
\midrule
\textbf{BART Variants} \\
\quad Full (1024) &52.83 &20.50 &50.14 \\
\quad Stride (4096) &54.29 &20.80 &51.35 \\
\quad LIN. (3072) &44.84 &13.87 &41.94 \\
\quad LSH (4096) &54.75 &21.36 &51.27 \\
\quad Sinkhorn (5120) &55.45 &21.45 &52.48\\
\textbf{BART HEPOS} \\
\quad LSH (7168) &55.00 &21.13 &51.67 \\
\quad Sinkhorn (10240) &\textbf{56.86} &22.62 &53.82 \\
\midrule
\ours & 56.77 & \textbf{23.25} & \textbf{53.90}
\\
\bottomrule
\end{tabular}
}
\caption{ROUGE scores on GovReport. For each baseline model, the number in parentheses is the maximum input length.
}
\label{tab:GovReport}
\end{table}

\subsection{Effects of Number of Stages}
\label{sec:stage}
\begin{figure}[!t]
    \centering
    \includegraphics[width=\linewidth]{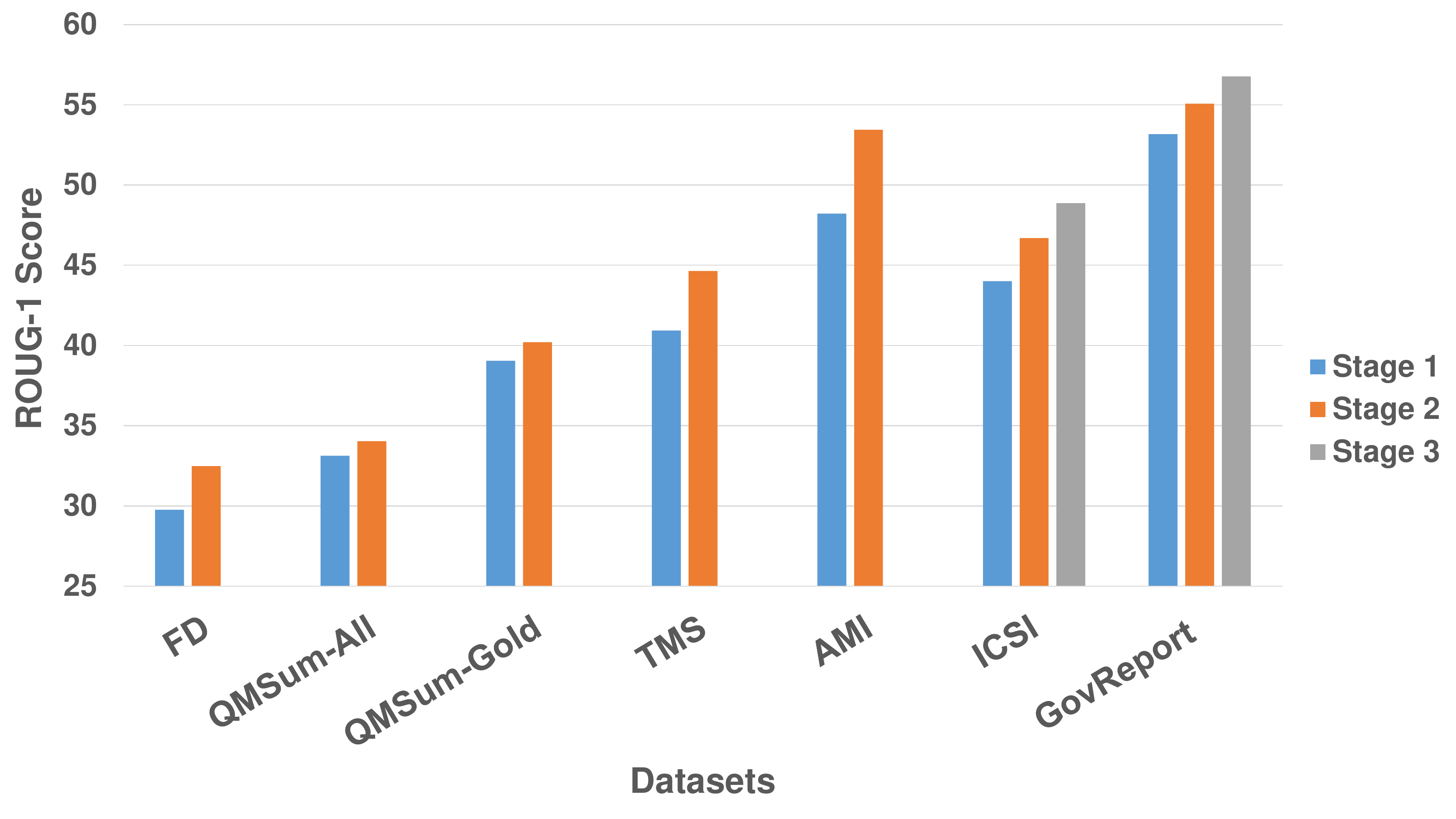}
    \caption{ROUGE-1 scores of various datasets at different stages. ICSI and GovReport have 3 stages, while the others have 2 stages. In all datasets, ROUGE-1 score increases with the increasing number of stages.
    }
    \label{fig:perf_stage}
\end{figure}

We also notice that the performance increases consistently when the number of stages goes up until the predefined number of stages.
Figure~\ref{fig:perf_stage} shows the ROUGE-1 scores of different tasks across stages. \textbf{Stage 1} indicates the model with only one coarse stage and no fine-grained stage. In this model, We directly use the first segment of the coarse summary as the output, i.e. $C_1^1$ of each sample. \textbf{Stage $\textbf{i}$} ($i>1$) model contains $i-1$ coarse stages and one fine-grained stage, the generated summary is from fine-grained summarization models, i.e. $F$.

Although stage 2 of \ours on the ICSI dataset has already outperformed the baselines, the scores can be further improved by adding one more coarse stage. In fact, on all datasets, increasing the number of stages leads to a performance gain. 
This gain can be explained as the following:
if the output of the current stage is longer than $K$ tokens, adding one more coarse stage will help since the model will receive more information from the source text compared with simply truncating them. On the contrary, if the input is smaller than $K$, there is no need to add more stages, because there is only one segment. 

\subsection{Improvements over Backbone Models}
\label{sec:improve}

\ours also boosts the performance of a backbone model by a large margin. As shown in Table~\ref{tab:backbone}, it improves the BART-large model by \textbf{6.87}, \textbf{3.89}, \textbf{6.78} ROUGE-1/2/L on AMI. This indicates the capability of \ours to boost the performance of a weak learner on long summarization tasks. In particular, when the backbone model is well pretrained on short input texts and performs well on short summarization tasks, \ours could greatly increase the capability of the backbone model to process and read long source texts.
Also, the backbone of \ours can be easily replaced by some other models, and models do not necessarily have to be identical at every stage. For example, one can try different learners such as T5 as the backbone model and replace the model in stage 1 with a dialogue-to-document model.

\begin{table}[t!]
\centering
\small

\begin{tabular}{@{}llrrr@{}}
\toprule
            &  & R1    & R-2   & R-L \\ \midrule
\multirow{2}{*}{AMI} & Backbone & 46.57 & 16.41 & 44.61 \\
                                     & \ours                       & \textbf{53.44} & \textbf{20.30} & \textbf{51.39}                \\ \midrule
\multirow{2}{*}{ICSI}  & Backbone                   & 39.91 & 9.98 & 38.17                  \\
                                     & \ours                       & \textbf{45.57} & \textbf{11.49} & \textbf{43.32}               \\\midrule
\multirow{2}{*}{QMSum-All}   & Backbone                   & 29.20 & 6.37 & 25.49                  \\
                                     & \ours                       & \textbf{34.03} & \textbf{9.28} & \textbf{29.48}               \\\midrule
\multirow{2}{*}{QMSum-Gold}    & Backbone                  & 32.18 & 8.48 & 28.56                        \\
                                     & \ours                     & \textbf{40.20} & \textbf{15.32} & \textbf{35.62}                \\ \bottomrule
\end{tabular}
\caption{Improvements of \ours over backbone BART models on AMI, ICSI, and QMSum datasets.} 
\label{tab:backbone}
\end{table}

\subsection{Generalizability over Backbone Models}
To demonstrate our framework can generalize to different backbone summarization models,
we replace the BART-large-cnn model in previous experiments with other neural summarization models including T5~\citep{2020t5} and PEGASUS~\citep{zhang2019pegasus} using Hugging Face.
Table~\ref{tab:replace} shows the ROUGE scores of three different models that are trained and evaluated on AMI. In all models, \ours improves the performance of backbone models by a large margin. For instance, although BART-base is a weaker summarizer compared with the BART-large model, the framework is still able to improve the ROUGE-1 score by 5.06. 

\begin{table}[t!]
\centering
\small
\resizebox{\linewidth}{!}{
\begin{tabular}{@{}llrrrr@{}}
\toprule
 &  & R-1   & R-2  & R-L  & Input\\ \midrule

\multirow{2}{*}{BART-base}  & Backbone & 41.54 & 13.80 & 38.75 & $1024$  \\
                                     & \ours  & 46.60 & 18.80 & 45.23   & $1024$ \\\midrule
\multirow{2}{*}{T5-large}   & Backbone & 47.81 & 16.06 & 45.77 & $512$ \\
                                     & \ours & 51.85 & 19.40 & 49.94   & $512$  \\\midrule
PEGASUS-   & Backbone & 46.37 & 16.21 & 44.75 & $1024$  \\
cnn\_dailymail  & \ours & 50.15 & 19.07 & 48.28 & $1024 $\\ \bottomrule
\end{tabular}}
\caption{ROUGE scores of different backbone models on AMI. For all backbone models with various maximum input lengths, ROUGE scores increase with the help of proposed framework. Input indicates the maximum number of tokens the model can take.
} 
\label{tab:replace}
\end{table}

\subsection{Ablations}
Table~\ref{tab:ablation} shows the ablation study results of \ours on the AMI test set. Removing stage 2 (using the first segment of the coarse summary $C^1_1$ as the generated summary) leads to a 5.23 ROUGE-1 score drop. Without data segmentation, the ROUGE-1 score decreases by 6.61 using the same fine-grained stage. Removing both stage 2 and target matching (use duplication algorithm instead) further decreases the performance. It even hurts the performance of the original BART model because the duplication of targets will introduce some biases towards the common part of the targets.

\begin{table}[t!]
\centering
\resizebox{.93\columnwidth}{!}{
\begin{tabular}{@{}lrrr@{}}
\toprule
          & R-1        & R-2       & R-L        \\ \midrule
\ours & \textbf{53.44} & \textbf{20.30} & \textbf{51.39} \\
\quad\quad - stage 2     & 48.21 & 18.59 & 46.46          \\
\quad\quad - data seg.      & 46.83          & 15.91          & 45.00          \\ 
\quad\quad - stage 2 \& tar. seg. & 46.24 & 16.03 & 44.45 \\
\quad\quad only BART & 46.57 & 16.41 & 44.61 \\
\bottomrule
\end{tabular}
}
\caption{Ablations on the test set of AMI. ``- data seg.'' indicates removing data segmentation (the same as cutoff at limitation), ``- tar. seg.'' indicates source segmentation paired with duplicated targets.
}
\label{tab:ablation}
\end{table}

\begin{table}[t!]
\resizebox{\linewidth}{!}{
\begin{tabular}{@{}lrrrrrr@{}}

\toprule
       & \multicolumn{3}{c}{AMI} & \multicolumn{3}{c}{ICSI} \\ 
       & Read.  & Conc.  & Cove. & Read.  & Conc.  & Cove.  \\\midrule
HMNet  &   3.93     & 4.05   &    4.15   &   3.21 &    3.33    &  3.84      \\
\ours  &   \textbf{4.45}    &   \textbf{4.13}     &   \textbf{4.23}     &  \textbf{4.12}      &     \textbf{3.55}  & \textbf{4.06}   \\\bottomrule
\end{tabular}
}

\caption{Human evaluation scores. Read. indicates \textit{Readability}, Conc. indicates \textit{Conciseness}, and Cove. indicates \textit{Coverage}.}
\label{tab:human}
\end{table}

\subsection{Human Evaluation}
We conduct a human evaluation to assess the following: \textit{Readability} takes into account word and grammatical error rate to evaluate how fluent the summary language is; \textit{Conciseness} measures how well the summary discards the redundant information; \textit{Coverage} measures how well the summary covers each part of the dialogue.

We compare the results of \ours and HMNet because HMNet is a baseline model with the good capability to read whole input. For each meeting in AMI and ICSI dataset, we ask 3 different annotators with English expertise to label the summaries.
Each annotator was asked to read the meeting transcript, gold summaries, and generated summaries using the SummVis~\citep{vig2021summvis} toolkit. They were asked to rate each summary from 1 to 5 (higher is better) for each metric. We also shuffle the summaries of two models to reduce the bias. 

Table~\ref{tab:human} shows that \ours achieves higher
scores in \textit{Readability}, \textit{Conciseness}, and \textit{Coverage} than HMNet in both AMI and ICSI dataset. Specifically, the \textit{Readability} of \ours greatly surpasses the baseline by around 0.5/1 point on AMI/ICSI dataset. This is because BART is well-pretrained and is able to generate more readable text and \ours successfully maintains this capability.

\begin{table}[!t]
\resizebox{\linewidth}{!}{
\begin{tabular}{@{}rrrrrr@{}}
\toprule
        & Avg. $|D^i|$    & Avg. $|C_i|$  & Comp. $R$ & $\hat{N}_{val}$ & $N$    \\ \midrule
Stage 1 & 7996.01 & 377.02 & - -     & 1.41  & 2 \\
Stage 2 & 3582.47 & 373.29 & 0.45 & 0.55  & 1 \\
Stage 3 & 1517.02 & 492.89 & 0.42 & -0.41 & 0 \\ \bottomrule
\end{tabular}}

\caption{Comparison of the empirical number of coarse stage $N$ and the corresponding estimation $\hat{N}_{val}$ on the GovReport test set. Avg. $|D^i|$ and Avg. $|C_i|$ are averaged number of tokens in source text and coarse segments of stage $i$ (Section \ref{sec:corstage}). Comp. $R$ is the compression rate $R$ of the stage.}
\label{tab:emp}
\end{table}

\subsection{Intermediate Result Analysis}
\label{sec:intermedi}

To gain more understanding of the multi-stage mechanism of \ours, we analyze the number of coarse stages and the compression rate through statistics of intermediate stages. 

\paragraph{Early Stopping of the Coarse Stage}
Although the ideal input of the final fine-grained stage should be shorter than $K$, the experiment results show that compressing input from 2$K$ to 1$K$ tokens usually hurts the performance of the model. This is probably because generating too many short segments which are hard to summarize confuses the model.

Thus, we increase the length of input to the final fine-grained stage from $K$ to $2K$ to prevent noises in the training set. The modified formula to estimate the number of coarse stages $\hat{N}$ is shown as follows (details in Appendix A).

    $$\hat{N}_{val} = \frac{1 +\log K - \log d_1}{\log c_1 - \log K}$$
    $$ \hat{N} = \lceil \hat{N}_{val} \rceil$$
    
\paragraph{Number of Coarse Stages}
To verify that our estimation $\hat{N}$ is close to the empirical number of coarse stages $N$, we use GovReport to compare the two as shown in Table \ref{tab:emp}.
We choose this dataset because it contains the most number of samples among all five datasets, with completely three coarse stages as well.

Table \ref{tab:emp} shows the empirical/estimated number of coarse stages. To clearly show the $\hat{N}$ value, we display the float number $\hat{N}_{val}$ as the estimated number, and $N$ as the empirical number of  ``remaining coarse stages'' (Table \ref{tab:datasets}). As can be seen, $N = \hat{N} = \lceil \hat{N}_{val} \rceil$ holds for all stages, meaning that the estimated $\hat{N}$ is capable of estimating the correct $N$ value. It is worth noting that, for stage 2 and stage 3, using this formula can also estimate ``how many additional coarse stage do we need''.

\paragraph{Compression Rate} We analyze the change of compression rate across different stages. In \ours, compression rate $R_i$ is defined as the averaged source length of stage $i$ divided by source length of stage $i-1$. As shown in Table \ref{tab:emp}, both compression rates in stage 2 and stage 3 of GovReport are around 0.4, this shows that the compression rate of \ours across different stages are stable, meaning that the number of segments will decrease to around 40\% of the previous stage steadily.

\begin{table}[!t]
\centering
\begin{tabular}{@{}lcc@{}}
\toprule
             & Transformers & \ours \\ \midrule
Time         & $O(n^2)$     &    $O(nK/(1-R))$  \\
Gen. Tokens  & $O(n)$       &    $O(n/(1-R))$   \\ \bottomrule
\end{tabular}
\caption{Time complexity of inference and the number of tokens generated during inference (Gen. Tokens) by comparing Transformers and \ours. $n$ is the number of tokens in the source text. $K$ is the maximum input length of the backbone model of \ours. $R$ is the averaged compression rate.}
\label{tab:time}
\end{table}

\subsection{Time Complexity} Table \ref{tab:time} shows the time cost of inferring one sample using vanilla transformer versus \ours. Although the \ours needs to generate more tokens due to multi-stage pipeline, \ours reduces the inference time from quadratic to lower, i.e., from $O(n^2)$ to $O(Cn)$, $C=K/(1-R)$. Regarding training the model, \ours also need to infer $O(n)$ additional tokens on the train/dev/test sets (details in Appendix B).

\section{Conclusion}
In this paper, we propose \ours, a simple, flexible, and effective framework for long dialogue and document summarization. It consists of multiple coarse stages and one fine-grained stage to iteratively compress the long source input. It enjoys the full power of backbone models while ensuring the full receptive field of the summarization model. We evaluate the model on various datasets and improve the baselines by a large margin.

\section*{Acknowledgement}
The authors would like to thank Tao Yu, Ming Zhong, Yixin Liu, and Asli Celikyilmaz for their valuable discussions. We also would like to thank the anonymous reviewers for their helpful comments. This work is supported in part by a grant from Microsoft Research.

\bibliography{anthology,custom}
\bibliographystyle{acl_natbib}
\input{appendix}

\end{document}

%% file: appendix.tex
\appendix

\begin{table*}[!t]
\resizebox{\textwidth}{!}{
\begin{tabular}{@{}ll@{}}
\toprule
         &  \makecell[c]{ICSI} \\ \midrule

\ours      & \makecell[l]{
The project manager opens the meeting by recapping the events of the \green{previous meeting}. The marketing expert presents the results of market \\ research , which shows that users want a fancy-looking remote control that is easy to use and has a \green{fancy look} and feel. The \green{user interface}\\ \green{designer} presents the user interface concept for the remote , which is based on the idea that a \green{remote} should be simple and user-friendly.\\ The industrial designer presents about the internal components of a remote control. The group discusses using kinetic \green{energy} to power the\\ device , using a simple battery for the \green{LCD screen} , and using an advanced chip for the advanced \green{chip}. The project manager closes the meeting\\ , telling the team members what their tasks will be for the next meeting. $\cdots$ The Marketing Expert will research how to produce a remote that\\ is technologically innovative. The User Interface Designer will look at how to make a remote \red{out of wood  or plastic with either a wooden}\\ \red{or plastic cover}. The Group will not work with teletext. There was a lack of information on the cost of components and materials. } \\ \midrule

Gold     & \makecell[l]{
The project manager opened the meeting and recapped the decisions made in the previous meeting. The marketing expert discussed his personal\\ preferences for the design of the remote and presented the results of trend-watching reports , which indicated that there is a need for products \\ which are fancy , innovative , easy to use , in dark colors , in recognizable shapes , and in a familiar material like wood. The user interface  \\ designer discussed the option to include speech recognition and which functions to include on the remote. The industrial designer discussed \\which options he preferred for the remote in terms of energy sources , casing , case supplements , buttons , and chips. The team then discussed \\and made decisions regarding energy sources , speech recognition , LCD screens , chips , case materials and colors, case shape and orientation , \\ and button orientation.$\cdots$ \gray{The case covers will be available in wood or plastic. The case will be single curved. Whether to use kinetic energy or}\\ \gray{a conventional battery with a docking station which recharges the remote. Whether to implement an LCD screen on the remote. Choosing} \\\gray{between an LCD screen or speech recognition. Using wood for the case.}} \\ \bottomrule
\end{tabular}
}
\caption{Sample output summary \ours on the ICSI dataset. Tokens marked in grey indicate the out-of-boundary contents of truncation models. Brown tokens are some topic words (manually selected) emerged in the gold summary. Tokens marked in red indicate the concepts of out-of-boundary text.} 
\label{tab:case}
\end{table*}

\section{Computing the Number of Stages}
With regard to text length, the source text of each stage needs to be compressed gradually to ensure that the summary with proper length can be generated in the final stage. Also, the compression level determines the required number of stages, which is a significant indicator of time cost. 

Suppose one sample of the source of stage $i$ contains $d_i = |D^i|$ words, while the source of next stage $D_{i+1}$ contains $d_{i+1} = |D^{i+1}|$ words. Also because the input of next stage is the coarse summary of current stage, $D^{i+1} = C^i$, thus $d_{i+1} = |D^{i+1}| = |C^i|$. The maximum input length of the model is $K$, $c_i = \sum_{j=0}^n |C_j^i|/n$ indicates the averaged number of tokens in the segmented predictions. $d_{i+1}$ can be expressed by the length of coarse summary which is the number of segment $\frac{d_i}{K}$ times the length of coarse segments $c_i$. 

In each stage, we have:
    $$ d_{i+1} = \frac{d_{i}}{K} \times c_{i} $$
By iterating this equation for $N$ time, the number of needed coarse stages $N$ for a dataset can be decided in this way:
    $$ d_1 \times \prod_{i=1}^{N}\frac{c_i}{K} \leq K $$
    
Empirically, $c_i$ are similar in different stages , thus we replace the production of $c_i$ with $c_1$ to the $N$, i.e.         $$c_1^N \approx \prod_{i=1}^{N}c_i$$

Thus, the estimation of $N$ value can be calculated as follows:
    $$ d_1 \times \frac{c_1^N}{K^N} \leq K $$ 
    $$ \hat{N} = \lceil \frac{\log K - \log d_1}{\log c_1 - \log K} \rceil$$

We also call $c_i/K$ the compression rate of stage $i$, denoted as $R_i$. For target matching, the compression rate of duplication segmentation is 1 and greedy segmentation is less than 0.5. So that target segmentation algorithm helps reduce number of coarse stages.

After using the early stopping of coarse stage, the estimation formula changes as follows:
$$ d_1 \times \frac{c_1^N}{K^N} \leq 2K $$ 
    $$ \hat{N} = \lceil \frac{1+\log K - \log d_1}{\log c_1 - \log K} \rceil$$
\section{Time Complexity}
Suppose the length of the input is $n$, by segmenting the source text into $n/K$ segments, the time cost of forwarding of one segment is $K^2$, thus the total time cost of stage 1 is $n/K \times K^2 = nK$. Then, in the next stage, the length of the source text is reduced to $nR$, thus the time complexity of stage 2 is $nKR$. We can list the total time cost by adding them together:

$$ T(n) = \sum_{i=0}^\infty nKR^i = \frac{nK}{1-R}$$

Similarly, in training phrase, stage 1 generates $O(n)$ tokens while stage 2 generates $O(nR)$ tokens for each sample in train/dev/test set. We can list the total generated tokens by adding them together:

$$ T(n) = \sum_{i=0}^\infty nR^i = \frac{n}{1-R}$$

 Thus the time cost of forwarding reduces. For instance, the inference time of SummScreen-TMS dataset reduces to $1024/(1-0.27)/6420.64 = 21.8\%$, and GovReport dataset reduces to  $1024/(1-0.43)/7890.46 = 22.8\%$ of original time cost, compared with $O(n^2)$ transformers. This shows the efficiency of \ours. On the other hand, since the training phrase needs to generate the target for each sample in the train/dev/test set, the training time of \ours also includes the additional generation of $O(\frac{n}{1-R})$ tokens for each sample in the dataset.

\section{Case Study}
Table \ref{tab:case} shows a concrete sample summary generated by \ours. It captures the topics of the source text and smoothly follows the outline of the gold summary. Also, \ours is able to evenly generate the information of the whole summary, including the last part of source text which is truncated in the standard BART-large models.